\documentclass[journal]{IEEEtran}
\usepackage{algorithm}
\usepackage{algorithmicx}
\usepackage{algpseudocode}
\usepackage{amsfonts,bm}
\usepackage{amsthm}
\usepackage{amsmath}
\usepackage{amssymb}
\usepackage{footnote}
\usepackage{perpage}
\usepackage{float}
\usepackage{stfloats}
\usepackage{mathrsfs}
\usepackage{graphicx}
\usepackage{multirow}
\usepackage{url}
\usepackage{cite}

\newcommand{\tabincell}[2]{\begin{tabular}{@{}#1@{}}#2\end{tabular}}

\begin{document}

\title{Efﬁciently Fusing Pretrained Acoustic and Linguistic Encoders for Low-resource Speech Recognition}

\author{{Cheng Yi, Shiyu Zhou, and Bo Xu, Member, IEEE}
\thanks{Cheng Yi is with the Institute of Automation, Chinese Academy of Sciences, China and University of Chinese Academy of Sciences, China (e-mail: yicheng2016@ia.ac.cn).}
\thanks{Shiyu Zhou is with the Institute of Automation, Chinese Academy of Sciences, China (e-mail: shiyuzhou@ia.ac.cn).}
\thanks{Bo Xu is with the Institute of Automation, Chinese Academy of Sciences, China (e-mail: xubo@ia.ac.cn).}}

\markboth{Journal of Class Files, Vol. *, No. *, January 2021}
{Shell \MakeLowercase{\textit{et al.}}: Bare Demo of IEEEtran.cls for IEEE Journals}
\maketitle

\begin{abstract}
    End-to-end models have achieved impressive results on the task of automatic speech recognition (ASR). 
    For low-resource ASR tasks, however, labeled data can hardly satisfy the demand of end-to-end models.
    Self-supervised acoustic pre-training has already shown its amazing ASR performance, while the transcription is still inadequate for language modeling in end-to-end models.
    In this work, we fuse a pre-trained acoustic encoder (wav2vec2.0) and a pre-trained linguistic encoder (BERT) into an end-to-end ASR model. The fused model only needs to learn the transfer from speech to language during fine-tuning on limited labeled data.
    The length of the two modalities is matched by a monotonic attention mechanism without additional parameters. Besides, a fully connected layer is introduced for the hidden mapping between modalities.
    We further propose a scheduled fine-tuning strategy to preserve and utilize the text context modeling ability of the pre-trained linguistic encoder.
    Experiments show our effective utilizing of pre-trained modules.
    Our model achieves better recognition performance on CALLHOME corpus (15 hours) than other end-to-end models.
\end{abstract}


\IEEEpeerreviewmaketitle

\section{Introduction}

\IEEEPARstart{P}{ipeline} methods decompose the task of automatic speech recognition (ASR) into three components to model: acoustics, pronunciation, and language\cite{zhou2017multilingual}. 
It can dramatically decrease the difficulty of ASR tasks, requiring much less labeled data to converge.
With self-supervised pre-trained acoustic model, the pipeline method can achieve impressive recognition accuracy with as few as 10 hours of transcribed speech \cite{baevski2020effectiveness,baevski2019vqwav2vec,baevski2020wav2vec,conneau2020unsupervised}.
However, it is criticized that the three components are combined by two fixed weights (pronunciation and language), which is inflexible \cite{pratap2018wav2letter++}.

On the contrary, end-to-end models integrate the three components into one and directly transform the input speech features to the output text. 
Among end-to-end models, the sequence-to-sequence (S2S) model is composed of an encoder and a decoder, which is the dominant structure \cite{dong2020cif,zhou2018multilingual,li2019speechtransformer,tran2020cross}.
The end-to-end modeling achieves better results than pipeline methods on most of public 
datasets \cite{li2019speechtransformer,dong2020comparison}.
Nevertheless, it requires at least hundreds of hours of transcribed speech for training. A deep neural network has an enormous parameter space to explore, which is hard to train with limited labeled data.

Pre-training can help the end-to-end model work well in the target ASR task on the low-resource condition \cite{hinton2012better}.
Supervised pre-training, also known as supervised transfer learning \cite{chung2018supervised}, uses the knowledge learned from other tasks and applies it to the target one \cite{zhou2018multilingual}.
However, this solution requires sufficient and domain-similar labeled data, which is hard to satisfy.
Another solution is to partly pretrain the end-to-end model with unlabeled data. 
For example, \cite{jiang2019improving} pre-trains the acoustic encoder of Transformer by masked predictive coding (MPC), and the model gets further improvement over a strong ASR baseline. 
Unlike the encoder, the decoder of the S2S model cannot be separately pre-trained since it is conditioned on the acoustic representation. In other words, it is difficult to guarantee the consistence between pre-training and fine-tuning for the decoder.

In this work, we abandon realizing linguistic pre-training for the S2S model. Instead, we turn to fuse a pre-trained acoustic encoder (wav2vec2.0) and a pre-trained linguistic encoder (BERT) into a single end-to-end ASR model. The fused model has separately exposed to adequate speech and text data, so that it only needs to learn the transfer from speech to language during fine-tuning with limited labeled data.
To bridge the length gap between speech and language modalities, a monotonic attention mechanism without additional parameters is applied. Besides, a fully connected layer is introduced for the mapping between hidden states of the two modalities. Our model works in the way of non-autoregressive (NAR) \cite{gu2017non} due to the absent of a well-defined decoder structure. NAR models have the speed advantage \cite{Higuchi2020} and can perform comparable results with autoregressive ones \cite{gu2017non}.
Different with self-training, acoustic representation is fed to the linguistic encoder during fine-tuning. 
The inconsistency can severely influence the representation ability of the linguistic encoder.
We help this module get along with the acoustic encoder by a scheduled fine-tuning strategy.

\section{Related Work}
A lot of works propose methods to leverage text data for the end-to-end model. Deep fusion \cite{8639038} and cold fusion \cite{Sriram2018,shan2019component} integrate a pre-trained auto-regressive language model (LM) into a S2S model. In this setting, the S2S model is randomly initialized and still needs a lot of labeled data for training.
Knowledge distillation \cite{8683602} requires a seed end-to-end ASR model, and it is not as convenient as the pretrain-and-finetune paradigm. 

\cite{tran2020cross} builds a S2S model with a pre-trained acoustic encoder and a multilingual linguistic decoder. The decoder is part of a S2S model (mBART \cite{liu2020multilingual}) pre-trained on text data. Although this model achieves great results in the task of speech-to-text translation, it is not verified in ASR tasks. Besides, this work does not deal with the  inconsistency we mentioned above.

\cite{Higuchi2020} takes advantage of an NAR decoder to revise the greedy CTC outputs from the encoder where low-conﬁdence tokens are masked. The decoder is required to predict the tokens corresponding to those mask tokens by taking both the unmasked context and the acoustic representation into account. \cite{chen2019listen} iteratively predicts masked tokens based on partial results. As mentioned above, however, these NAR decoders cannot be separately pre-trained since they rely on the acoustic representation.

\section{Methodology}
\label{sec:model}
    \begin{figure*}[th]
    \centering
    \includegraphics[width=1.0\linewidth]{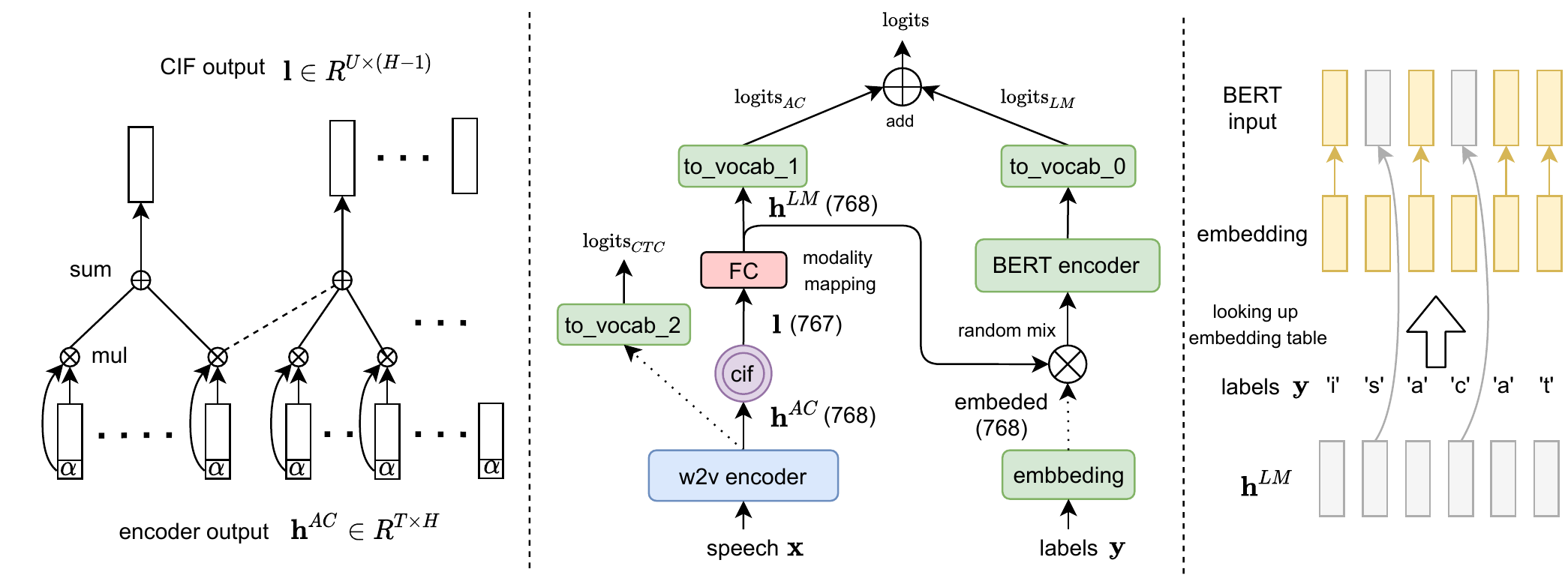}
    \caption{The structure of w2v-cif-bert. On the left part, a variant CIF mechanism converts $\bm h^{AC}$ to $\bm l$ without any additional module; on the right part, $\bm h^{LM}$ is not directly fed into BERT but mixed with embedded labels in advance during training. 
    Modules connected by dot lines are ignored during inference. Numbers in ``()'' indicate the size of hidden in the model.}
    \label{fig:model}
    \end{figure*}

We propose an innovative end-to-end model called \textbf{w2v-cif-bert}, which consists of wav2vec2.0 (pre-trained on speech corpus), BERT (pre-trained on text corpus), and CIF mechanism to connect the above two modules. The detailed realization of our model is demonstrated in Fig. \ref{fig:model}. It is worth noting that only the fully connection in the middle (marked as red) does not participate in any pre-training.

\subsection{Acoustic Encoder}
\label{ssec:wav2vec}
We choose wav2vec2.0 as the acoustic encoder since it has been well verified \cite{baevski2020wav2vec,yi2019ectc,tran2020cross}.
Wav2vec2.0 is a pre-trained encoder that converts raw speech signals into acoustic representation \cite{baevski2020wav2vec}. During pre-training, it masks the speech input on the latent space and solves a contrastive task.
Wav2vec2.0 can outperform previous semi-supervised methods simply through fine-tuning on transcribed speech with CTC criterion \cite{graves2006connectionist}. 

Wav2vec2.0 is composed of a feature encoder and a context network. The feature encoder outputs with a stride of about 20ms between each frame, and a receptive field of 25ms of audio. The context network is a stack of self-attention blocks for context modeling \cite{vaswani2017attention}. In this work, wav2vec2.0 is used to convert speech $\bm x$ to acoustic representation $\bm h^{AC}$ (w2v encoder) in our model, which is colored blue in Fig. \ref{fig:model}.

\subsection{Modality Adaptation}
\label{ssec:cif}
It is normal to apply global attention to connect acoustic and language representation \cite{vaswani2017attention}. However, this mechanism is to blame for the poor generalization of text length \cite{dong2020comparison}, which is worse under sample scarcity.
Instead, we use continuous integrate-and-fire (CIF) mechanism \cite{dong2020cif} to bridge the discrepant sequence lengths. CIF constrains a monotonic alignment between the acoustic and linguistic representation, and the reasonable assumption drastically decreases the difficulty of learning the alignment.

In the original work \cite{dong2020cif}, CIF uses a local convolution module to assign the attention value to each input frame. 
To avoid introducing additional parameters, the last dimension of the $\bm h_{AC}$ is regarded as the raw scalar attention value (before sigmoid operation), as demonstrated in the left part of Fig. \ref{fig:model}.
The normalized attention value $\alpha_t$ are accumulated along the time dimension $T$ and a linguistic representation $\bm l_u$ outputs whenever the accumulated $\alpha_t$ surpasses 1.0.
During training, the sum of attention values for one input sample is resized to the number of output tokens $\bm y$ ($n^* = ||\bm y||$). 

The formalized operations in CIF are:

    \begin{align}
        \alpha_t &= \text{sigmoid}(\bm h^{AC}_t[-1]) \label{eq:sigmoid} \\
        \alpha'_t &= \text{resize}(\alpha_t |\bm \alpha_T, n^*) \label{eq:resize} \\
        \bm l_u &= \sum_t\alpha'_t * \bm h^{AC}_t[:-1]
    \end{align}
, where $\bm h^{AC}$ represents acoustic vectors with length of $T$, $\bm l$ represents accumulated acoustic vectors with length of $U$. 


CIF introduces a quantity loss to supervise the encoder generating the correct number of final modeling units:
    \begin{align}
        \hat{n} &= \sum_t^T \alpha_t \\
        L_{qua} &= ||n^* - \hat{n}||_2
    \end{align}
, where $\hat{n}$ represents the predicted decoding length.
During inference, we add an extra rounding operation on $\hat{n}$ to simulate $n^*$ between Eq. \ref{eq:sigmoid} and \ref{eq:resize}.

Based on the matched sequence length, the accumulated acoustic vector $\bm l$ is mapped into the linguistic vector $\bm h^{LM}$ by a randomly initialized fully connected layer (FC), realizing the modality adaptation.

\subsection{Linguistic Encoder}
\label{ssec:lm}
We choose BERT as the linguistic encoder in our model.
BERT is a masked LM, applying the \textit{mask-predict} criterion for self-training and utilizes both left and right context on a huge amount of text data \cite{devlin2019bert}.
BERT has empirically shown impressive performance on various NLP tasks \cite{devlin2019bert,lan2019albert,yang2019xlnet}.

BERT is composed of three modules: an embedding table (embedding) to convert tokens to hidden vectors, a final fully connection layer (to\_vocab\_0) to convert hidden vectors to an output softmax over the vocabulary, and a Transformer encoder (BERT encoder) for bidirectional context modeling.
These modules are colored green in Fig. \ref{fig:model}.
Pre-trained BERT can compensate for the lack of text data on low-resource ASR tasks. Specifically, the prior output distribution can accelerate the convergence, and BERT provides stronger linguistic context. 

\subsection{Additional Connections}
\label{ssec:connection}
We add two additional connections after stacking the three modules.
Firstly, a high-way directly connects the acoustic to the final output. Secondly, an auxiliary CTC supervision is attached to the acoustic encoder ($L_{ctc}$).
Both of the two connections can make the encoder affected by the target supervision more effectively \cite{dong2020cif,kim2017joint}. 
We use BERT's final fully connected layer (to\_vocab\_0) to initialize the new ones (to\_vocab\_1 and to\_vocab\_2).
The final output of our model is:
    \begin{equation}
        \text{logits} = \lambda_{AC} * \text{logits}_{AC} + \lambda_{LM} * \text{logits}_{LM}
    \end{equation}
, where $\lambda_{AC}$ and $\lambda_{LM}$ are the weights for the output from the acoustic encoder and the linguistic encoder. The final output is supervised by the cross-entropy criterion ($L_{ce}$) \cite{vaswani2017attention}.

The final loss during fine-tuning over labeled speech is:
    \begin{equation}
        L = L_{ce}  + \mu_1 * L_{qua} + \mu_2 * L_{ctc}
    \end{equation}
, where $\mu_1$ and $\mu_2$ are the weights for these losses respectively.

\subsection{Scheduled Modality Fusion}
\label{ssec:sf}
We notice a distinct mismatch between BERT as a text feature extractor during pre-training and as a linguistic encoder in the ASR model during fine-tuning. 
BERT cannot process $\bm h^{LM}$ according to its pre-trained knowledge of text processing. 
It needs to greatly adjust the parameters on the bottom, which is significantly different from fine-tuning on NLP tasks.
Worst of all, BERT's parameters cannot transfer from the bottom up with a minimum cost. On the contrary, BERT will undergo a massive top-down change following the broadcast of gradients. 

In response to the above mismatch, we randomly replace $\bm h^{LM}$ with embedded target tokens $\bm y$ along linguistic length $U$ with a scheduled \textit{gold rate} $p\in [0,1]$, as demonstrated in the right part of Fig. \ref{fig:model}.
$p$ will decrease during fine-tuning for BERT to get rid of the dependency on $\bm y$.
At the beginning of fine-tuning, BERT cannot understand the frames from $\bm h^{LM}$ and views them as \textit{masked} input. BERT mainly utilizes the gold context (label embedding vectors) to predict. Due to the consistency with pre-training, BERT can fastly converge to high predicting accuracy. Along with the fine-tuning and the decrease of $p$, BERT gradually grasps the meaning of $\bm h^{LM}$ and predicts more accurately. 
During inference, $p$ is set to 0 and BERT needs to predict with pure $\bm h^{LM}$.


\section{Experiments}
\label{sec:experiments}

\subsection{Datasets and Experimental Settings}
\label{sec:setting}
We focus on low-resource ASR and mainly experiment on CALLHOME corpus \cite{zhou2017multilingual}.
CALLHOME is a multilingual corpus with less than 20 hours of transcribed speech for each language.
In this work, we use CALLHOME Mandarin (MA, LDC96S15) and English (EN, LDC97S20).
MA has 23,915 transcribed utterances (15.6h) for training and 3,021 for testing. EN has 21,194 transcribed utterances (14.9h) for training and 2,840 for testing. 
To compare with more works, we also test our model on a relatively large and popular corpus:  HKUST \cite{liu2006hkust}.
HKUST corpus (LDC2005S15, LDC2005T32) consists of a training set and a development set, which adds up to about 178 hours.
Both of them are telephone conversational speech corpus, which is much more realistic and harder than Librispeech \cite{baevski2020wav2vec}.

We use the open source wav2vec2.0 \footnote{\url{https://dl.fbaipublicfiles.com/fairseq/wav2vec/wav2vec_small.pt}} as the encoder, 
bert-base-uncase \footnote{\url{https://storage.googleapis.com/bert_models/2020_02_20/uncased_L-12_H-768_A-12.zip}} as the linguistic encoder for English and bert-base-chinese \footnote{\url{https://storage.googleapis.com/bert_models/2018_11_03/chinese_L-12_H-768_A-12.zip}} as the linguistic encoder for Chinese. We are free from pre-processing the transcripts since the built-in tokenizers of BERTs can automatically generate the modeling units.
All the code and experiments are implemented using \textit{fairseq} \cite{ott2019fairseq}. We reserve most of the training settings in wav2vec2.0 fine-tuning demonstration \footnote{\url{https://github.com/pytorch/fairseq/tree/master/examples/wav2vec}}. We optimize with Adam, warming up the learning rate for 8000 steps to a peak of $4\times10^{-5}$, holding 42000 steps and then exponential decay it. we only use a single GPU (TITAN Xp) for each experiment. Considering the NAR property, our model simply uses the greedy search to generate final results.

\subsection{Overall Results}
\label{ssec:compare}
In this section, we compare our model with other end-to-end modeling works. 
Transformer \cite{zhou2018multilingual} conducts experiments on low-resource tasks (MA and EN) by supervised transfer learning.
w2v-ctc \cite{yi2020applying} also applies pre-trained wav2vec2.0 as encoder and
adds a randomly initialized linear projection on top of the encoder. w2v-ctc is optimized by minimizing a CTC loss, and it is one of the most concise end-to-end models \cite{graves2014towards}.

In this work, we reproduce w2v-seq2seq \cite{yi2020applying}, which is composed of pre-trained wav2vec2.0 and Transformer decoder (with 1 or 4 blocks, randomly initialized) along with cross-attention \cite{vaswani2017attention}.
w2v-seq2seq models apply the same modeling units (character for Chinese and subword for English) as \cite{yi2020applying}.
We further implement \textit{cold fusion} for w2v-seq2seq with pre-trained Transformer LM (6 blocks) on a private Chinese text corpus (200M samples). w2v-seq2seq models decode through the beam search with a size of 50. All of these models are trained under the same setups as w2v-cif-bert.

    \begin{table}[th]
        \centering
        \caption{Comparison on CALLHOME and HKUST. Performance is CER (\%) for MA and HKUST; WER (\%) for EN.}
        \begin{tabular}{ l c c c}
        \hline
        \hline
        \textbf{Model} & MA(15h) & EN(15h) & HKUST(150h) \\
        \hline
        CIF \cite{dong2020cif} & - & - & 23.09 \\
        Transformer + MPC \cite{jiang2019improving} & - & - & \textbf{21.70} \\
        Transformer \cite{zhou2018multilingual,zhou2018comparison} & 37.62 & 33.77 & 26.60 \\
        w2v-ctc \cite{yi2020applying} & 36.06 & 24.93 & 23.80 \\
        \hline
        w2v-seq2seq \\
        \quad decoder with 1 blocks & 39.81 & 26.18 & 24.06 \\
        \qquad + cold fusion & 37.90 & - & 24.02 \\ 
        \quad decoder with 4 blocks & 54.82 & 47.66 & 25.73 \\
        \qquad + cold fusion & - & - & 25.46 \\ 
        \hline
        w2v-cif-bert & \textbf{32.93} & \textbf{23.79} & 22.92 \\
        \hline
        \hline
        \end{tabular}
    \label{tab:compare1}
    \end{table}

As demonstrated in Table \ref{tab:compare1}, our model achieves best results on low-resource tasks, showing the promising direction to fuse pre-trained acoustic and linguistic modules.
On the relatively abundant labeled ASR task, our model still achieves comparable performance as the SOTA. Considering the MPC training \cite{jiang2019improving} utilizes a rather large speech data (10,000 hours) that is similar to the target task, this benchmark is hard to approach.

w2v-seq2seq models are inferior to w2v-ctc, even with cold-fusion. We think it is the random-initialized decoder that causes the low performance under the low-resource condition.
Compared with cold-fusion, our method can make better use of pre-trained LM.

\subsection{Ablation Study}
\label{ssec:ablations}
We explore different structures and hyper-parameters of w2v-cif-bert by ablations to find a reasonable setting. Some connections in Fig. \ref{fig:model} are shut off by setting corresponding weights to 0. Ablations are conducted on CALLHOME-MA.
    
    \begin{table}[h]
        \centering
        \caption{Ablations on structure of w2v-cif-bert over CALLHOME MA. Performance is CER (\%) on test set.}
        \begin{tabular}{ l l l }
        \hline
        \hline
        \textbf{Description} & \textbf{Settings} & \textbf{CER} \\
        \hline
        w2v-cif-bert & \tabincell{l}{
        $\mu_1=0.2$ \\ 
        $\mu_2=1.0$ \\
        $\lambda_{LM}=0.2$ \\
        $\lambda_{AC}=1.0$ \\
        no sharing to\_vocab \\
        $0.9 \rightarrow 0.2 / 4000$ \\
        $TH=0.8$} & 
        \textbf{32.93}\\
        \hline
        \multirow{2}{*}{(1) quantity loss} & 
        $\mu_1=0$ & 154.7 \\ 
        & $\mu_1=0.5$ & 33.05 \\
        \hline
        \multirow{2}{*}{(2) CTC loss} &
        $\mu_2=0$ & 35.77 \\ 
        & $\mu_2=2.0$ & 33.01 \\
        \hline
        \multirow{2}{*}{(3) LM weight}
        & $\lambda_{LM}=0.0$ & 36.43 \\
        & $\lambda_{LM}=0.4$ & 33.22 \\
        \hline
        (4) acoustic weight & $\lambda_{AC}=0.0$ & 99.69 \\
        \hline
        \multirow{3}{*}{(5) share to\_vocab} 
        &share 0,1 & 33.00 \\
        &share 0,2 & 78.76 \\
        &share 1,2 & 35.10 \\
        \hline
        \multirow{5}{*}{(6) gold rate schedule} 
        & $0.9 \rightarrow 0.2 / 8000$ & 34.13 \\
        & $0.9 \rightarrow 0.2 / 2000$ & 33.37 \\
        & $0.9 \rightarrow 0.0 / 4000$ & 34.92 \\
        & $0.2 \rightarrow 0.2 / \infty$ & 33.26 \\
        & $0.0 \rightarrow 0.0 / \infty$ & 35.95 \\
        \hline
        \multirow{2}{*}{(7) confidence threshold} 
        & $TH=1.0$ & 33.37 \\
        & $TH=0.6$ & 35.51 \\
        \hline
        \hline
        \end{tabular}
    \label{tab:ablation}
    \end{table}

Results are listed in Table \ref{tab:ablation}. We get the following conclusions according to the corresponding ablations:
(1) The quantity loss is inevitable for the CIF mechanism since a proper alignment between the acoustic and linguistic representation is hard to learn through other supervisions;
(2) Adding the auxiliary CTC criterion greatly matters. It can help the encoder to learn the alignment;
(3) BERT as linguistic encoder makes an impressive contribution to the performance, showing our effective utilization of pre-trained masked LM;
(4) The acoustic high-way connection is inevitable for the performance convergence of our model, which plays a similar role as the CTC criterion;
(5) Using three separate-updating to\_vocab modules is the best option.

During fine-tuning, we apply a trick of mixing $\bm h^{LM}$ with embedded labels at a scheduled sampling rate $p$. we explore different schedules of $p$ in (6). ``$\rightarrow$'' indicates the range of $p$, and the number after ``/'' is the decreasing step.
Directly fine-tuning without any mixing ($0.0 \rightarrow 0.0 / \infty$) achieves a rather poor performance, demonstrating the necessity of our proposed schedule fusing trick. 
A reasonable schedule of gold rate $p$ during fine-tuning is decreasing from a high value to a low one. Keeping $p>0$ is another key point. We explain that it can make BERT work better by preserving some anchor tokens.

During inference, we also add some anchor tokens according to the conﬁdence of the acoustic output $\text{logits}_{AC}$, which is similar to the iterative decoding of NAR \cite{Higuchi2020}. Tokens are determined when their post-probabilities surpass a customized threshold $TH$ ($TH=1.0$ means no anchor tokens). Then the embedding vectors of these tokens are mixed with $\bm h^{LM}$.
As we can see in (7), a proper confidence threshold, $TH=0.8$ in our best result, contributes to the better performance during inference.

\vfill\pagebreak
\bibliographystyle{IEEEtran}
\bibliography{mybib}

\end{document}